\documentclass[10pt,twocolumn,letterpaper]{article}
\usepackage{cvpr} 

\definecolor{cvprblue}{rgb}{0.21,0.49,0.74}
\usepackage[pagebackref,breaklinks,colorlinks,allcolors=cvprblue]{hyperref}
\usepackage{multirow}

\def\paperID{37290} 
\def\confName{CVPR}
\def\confYear{2026}

\title{Same Attention, Different Truths: Put Logit-Lens over Visual Attention to\\ Detect and Mitigate LVLM Object Hallucination}

\author{
Zichuan Wang$^{1,2}$ \quad Songlin Yang$^{3}$ \quad Bo Peng$^{2}$\thanks{Corresponding author.} \quad Zhenchen Tang$^{1,2}$ \quad Yang Li$^{2}$ \\ Beibei Dong$^{2}$ \quad Jing Dong$^{2}$\\
$^{1}$School of Artificial Intelligence, University of Chinese Academy of Sciences\\
$^{2}$New Laboratory of Pattern Recognition, Institute of Automation, Chinese Academy of Science\\
$^{3}$Hong Kong University of Science and Technology\\
{\tt\small
\{wangzichuan2024,tangzhenchen2024,liyang2022,dongbeibei2022\}@ia.ac.cn} \\
{\tt\small syangds@connect.ust.hk,\{bo.peng,jdong\}@nlpr.ia.ac.cn}}
\begin{document}
\maketitle
\begin{abstract}
Large Vision-Language Models (LVLMs) often suffer from object hallucination, generating objects that are absent from the image. Prior work largely attributes this to insufficient visual attention. However, we find that both real and hallucinated objects receive equally strong visual attention in model’s mid-to-late layers, suggesting that the key issue may not be how much the model attends, but \textbf{what it attends to and why}. To this end, we decode the visual features of high-attention regions using Logit Lens, and observe that regions corresponding to real objects can be correctly decoded to the target object tokens, whereas those for hallucinated objects cannot. Building on this, we identify two hallucination mechanisms: \textbf{(i) visual uncertainty}, triggered by semantically similar or confusable regions, masking these regions eliminates the hallucination. \textbf{(ii) contextual prior}, triggered by strong co-occurrence priors, even when the initially attended region is masked, the hallucination persists and attention drifts to other regions. Based on these findings, we propose a simple yet effective training-free \textbf{Detect–Mitigate framework} comprising a Logit-Lens Consistency Check to detect hallucination and targeted remedies: High-Attention Regions Masking (HARM) for visual uncertainty hallucination, and Visual Evidence Enhanced Decoding (VEED) for contextual prior hallucination. Our approach achieves state-of-the-art results on multiple hallucination benchmarks. Code will be available.\footnote{\url{https://github.com/wzczc/SADT}}


\end{abstract}
\begin{figure}[t]
  \centering
  \includegraphics[width=\linewidth]{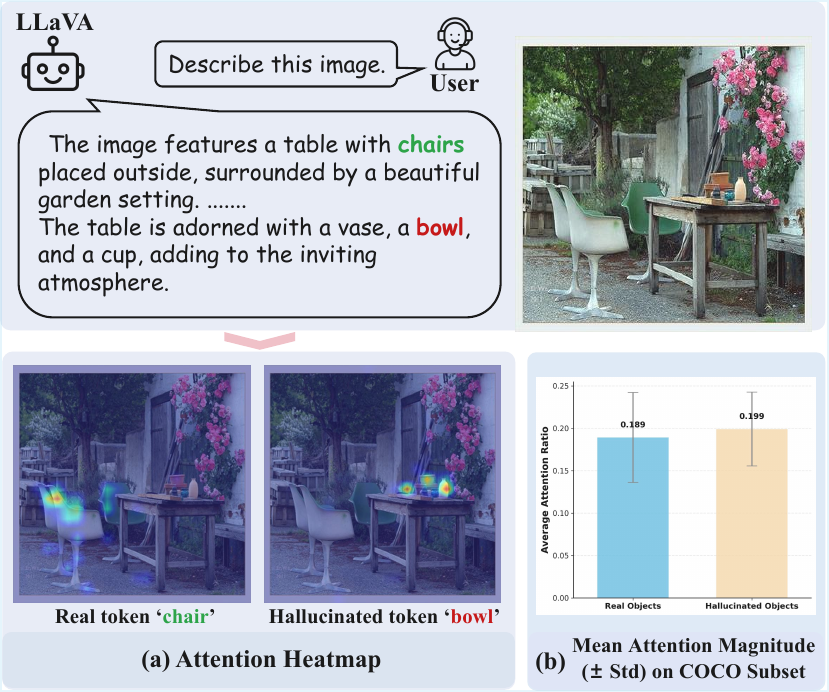}
  \caption{Different from prior work that attributes object hallucination to insufficient visual attention, we find that real and hallucinated objects often have the same attention magnitudes, as shown qualitatively in \textbf{(a)} and quantitatively in \textbf{(b)}, reflecting ``Same Attention, Different Truths". 
  }
  \label{fig:same_attention}
\end{figure}

\vspace{-6mm}
\section{Introduction}
\label{sec:intro}

Large Vision-Language Models (LVLMs) have made significant strides in recent years, demonstrating impressive capabilities on cross-modal tasks~\cite{llava,qwen-vl,internvl,blip2,instructblip,mvbench}. However, they still suffer from object hallucination~\cite{evaluating_object_hallucination,survey,survey1}, i.e., generating objects that are not present in the image, undermining their reliability and practical use.  

A prevailing view attributes hallucination to insufficient visual attention, either because textual priors dominate the multimodal interaction and suppress visual focus~\cite{what_in,ict,eyes,evaluating_object_hallucination,overrules}, or because improper processing of visual information leads to inadequate attention to salient regions~\cite{devils,agla,see}. Accordingly, many methods attempt to amplify or redistribute attention~\cite{pai,damro,clearsight,devils,see}, or inject extra visual guidance during training/inference~\cite{with_visual_supervision,data-augmented,vcd,classifier-free_guidance}. However, our experiments reveal that this explanation is incomplete: \textit{the key issue may not be how much the model attends to the image, but what it attends to and why}. To support this claim, we first examine \textbf{how much} the model attends. By quantifying the attention strength associated with object tokens, we find that visual attention consistently peaks in the mid-to-late layers, which we define as the ``image attention stage''. Both real and hallucinated object tokens exhibit similar attention magnitudes and focus on specific regions in this stage, as shown in Fig.~\ref{fig:same_attention}. Thus, the occurrence of hallucination cannot be explained simply by insufficient visual attention.  

Next, to understand \textbf{what} the model attends to, we analyze the semantics behind these high-attention regions using the Logit Lens~\cite{logitlens}, which uses the model’s textual decoding head to translate visual representations into text tokens. We observe that regions corresponding to real object tokens can be accurately decoded into their target tokens, whereas those corresponding to hallucinated object tokens fail to produce semantically consistent outputs. This reveals a semantic misalignment between visual semantics and output tokens within high-attention regions, showing that hallucination is typically accompanied by inconsistent semantic decoding, rather than by the lack of visual attention.  

To further probe \textbf{why} the model attends to these regions when hallucination occurs, we mask these regions and examine how the model’s output changes. This intervention reveals two distinct hallucination mechanisms. \textbf{(1) Visual Uncertainty Hallucination}: triggered by confusing or semantically related regions (e.g., a fuzzy ceramic-like area mistaken for a ``bowl''). Masking these high-attention regions directly eliminates the hallucination. This shows that the model cannot extract precise semantics from highly uncertain visual evidence, leading to decoding inconsistency. \textbf{(2) Contextual Prior Hallucination}: dominated by co-occurrence priors (e.g., generating ``microwave'' in a description of a kitchen even though it is absent). In this case, masking the high-attention regions does not remove the hallucination, and attention shifts to other regions. Although these regions can be correctly decoded into their target tokens, they are not selected in the final output due to the suppression by strong priors. Here, the attention mechanism behaves more like a ``routine'': the model has learned that it must attend to some region before generating an object and thus anchors attention merely to satisfy this requirement.

Based on these findings, we propose a simple yet effective training-free \textbf{Detect–Mitigate framework} to precisely locate and reduce hallucinations. First, we propose the Logit-Lens Consistency Check to detect hallucinated tokens by comparing whether the semantic decoding of high-attention regions in the mid-to-late layers is consistent with the token being generated. Then, we propose corresponding mitigation methods for each category. For the visual uncertainty cases, we perform simple High-Attention Regions Masking (HARM) to remove unreliable visual evidence. For the contextual prior cases, we propose Visual Evidence Enhanced Decoding (VEED). Here, the high-attention regions can still be decoded to the correct tokens via Logit Lens, but are suppressed during generation due to the strong priors. We therefore inject the true visual evidence from these regions into the generation process, enhancing the contribution of image-grounded semantics and suppressing contextual priors. Finally, we apply our strategy to multiple LVLMs on several hallucination benchmarks, achieving SOTA results, demonstrating its effectiveness and generality.
In summary, our contributions are as follows:
\begin{itemize}
    \item We reveal the ``same attention, different truth'' phenomenon, showing the key of object hallucination is the semantic inconsistency between high-attention visual features and the text output, not the attention strength. Based on this, we categorize object hallucination into two core mechanisms: visual uncertainty and contextual prior.
    \item We propose a simple yet effective training-free ``Detect-Mitigate'' framework, which includes: (1) the Logit-Lens Consistency Check for detection and (2) two mitigation strategies targeting each mechanism: High-Attention Regions Masking and Visual Evidence Enhanced Decoding.
    \item Our method achieves SOTA performance on multiple object hallucination benchmarks across several LVLMs, proving its effectiveness and generalizability.
\end{itemize}

\section{Related Work}
\label{sec:related}
\subsection{Large Vision-Language Models}
Recent years have witnessed remarkable progress in Large Vision-Language Models (LVLMs), which integrate language models with visual encoders to perform unified multimodal capacity.  
Early work such as CLIP~\cite{clip} established large-scale vision–language pretraining via contrastive image–text alignment. Subsequent models, including BLIP-2~\cite{blip2} and Flamingo~\cite{flamingo}, further bridged modality gaps via learnable adapters or cross-attention mechanisms. More recently, instruction-tuned LVLMs such as LLaVA~\cite{llava,llava1.5}, Shikra~\cite{shikra},  mPLUG-Owl2~\cite{mplug}, Qwen-VL~\cite{qwen-vl} and InternVL~\cite{internvl} have demonstrated impressive open-ended generation and reasoning capabilities. Despite these advances, LVLMs still frequently hallucinate objects, i.e., describe objects that are absent from the image, which undermines their reliability in practical applications.

\subsection{Object Hallucination in LVLMs}
Object hallucination is a critical challenge for LVLMs. Existing mitigation methods fall into two categories. The first is training-stage methods, such as introducing additional datasets~\cite{with_visual_supervision,mitigating_instruction}, incorporating new training objectives~\cite{Hallucination_contrastive,hallucination-induced_optimization}, and reinforcement learning methods~\cite{on-policy-dpo,ha-dpo,rlhf-v}. However, they are costly in terms of time and resources. The second is inference-stage methods, including contrastive decoding~\cite{cd,vcd}, guided decoding~\cite{classifier-free_guidance,halc}, and visual enhancement~\cite{agla,clearsight}. Recently, a growing body of work has begun to investigate the internal mechanisms of hallucination at inference time, with many studies attributing it to insufficient visual attention. ~\cite{pai,vcd,overrules,clearsight} find that textual priors often dominate multimodal interaction. In parallel, ~\cite{see,agla,eyes, devils} show that improper processing of visual information also results in weak visual attention. Building on these, ~\cite{see,clearsight,devils} propose reallocating or amplifying attention to strengthen the visual pathway, while ~\cite{pai,vcd,deco} leverage visually guided to counteract language priors.
Our work follows the attention-based analysis but takes a step further. We argue that the ``insufficient visual attention" explanation is incomplete, as our findings show real and hallucinated objects can receive comparable attention magnitudes, while the semantic consistency between the attended regions and the textual output may be the key.
\section{Object Hallucination Analysis}
\label{sec:object}
In this section, we analyze the mechanisms underlying object hallucination in LVLMs. We begin with the necessary preliminaries, then compare the attention patterns of hallucinated and non-hallucinated objects and use the Logit Lens to identify their semantic differences. Finally, we conduct intervention experiments that categorize object hallucination into two distinct types.

\subsection{Preliminary}
\textbf{LVLM Generation Process.} 
LVLMs typically consists of a vision encoder, a cross-modal projector and a LLM. Given an image \(I\) and a textual prefix \(Q_{<t}=\{q_1,\dots,q_{t-1}\}\), the vision encoder maps \(I\) to image tokens \(V=\{v_1,\dots,v_N\}\), which the projector \(M\) transforms into \(\tilde v_i=M(v_i)\). These tokens are then concatenated with system tokens \(S=\{s_1,\dots,s_M\}\) and the text prefix to form the input sequence \(X=[\,s_1,\dots,s_M,\tilde v_1,\dots,\tilde v_N,\,q_1,\dots,q_{t-1}]\), which is fed into a multi-layer transformer decoder with causal masking. At generation step \(t\), the output head \(W_U\) maps the hidden state \(h_t^{(L)}\) of the last layer $L$ to the next-token distribution:
\begin{equation}
    p(x_t\mid I,X_{<t})=\operatorname{softmax}(W_U h_t^{(L)}) .
\end{equation}

\textbf{Attention Mechanism.} In LVLMs, attention transfers information between tokens. The autoregressive transformer uses causal attention: at step \(t\), the query can only attend to tokens at positions \(j \le t\), while positions \(j>t\) are masked. Let \(q_t\) be the query, and \(k_j\) be the keys. The mask \(m_{tj}\) is defined as \(m_{tj} = 0\) when \(j \le t\) and \(-\infty\) otherwise. The attention weight between tokens are then computed as:
\begin{equation}
    \alpha_{tj} = \mathrm{softmax} \left( \frac{q_t^\top k_j}{\sqrt{d_k}} + m_{tj} \right).
\end{equation}

\textbf{Logit Lens.} Logit Lens is an interpretability method that decodes the hidden states of intermediate layers in LLM into vocabulary space, revealing how information evolves across layers. Given a hidden state \( h_t^{(l)} \), the output head \( W_U \) produces a probability distribution over the vocabulary:
\begin{equation}
p_{\text{lens}}^{(l)}(\cdot \mid h_t^{(l)}) = \mathrm{softmax}\left( W_U \, h_t^{(l)} \right).
\end{equation}
In this work, we apply it to the hidden states of image tokens to interpret how the model processes visual information at different layers.

\subsection{Same Attention: Attention Magnitude Is Not to Blame}
\label{subsec:same attention}
To analyze whether attention magnitude play a role in generating hallucinated object tokens,
we adopt LLaVA-1.5-7B~\cite{llava1.5} as an example, randomly sampling 500 images from the COCO 2014 validation set~\cite{coco} and use the prompt ``Describe this image.'' for generation. For each generated token, we collect its attention distribution data across all the layers.
\begin{figure}[t]
  \centering
  \includegraphics[width=\linewidth]{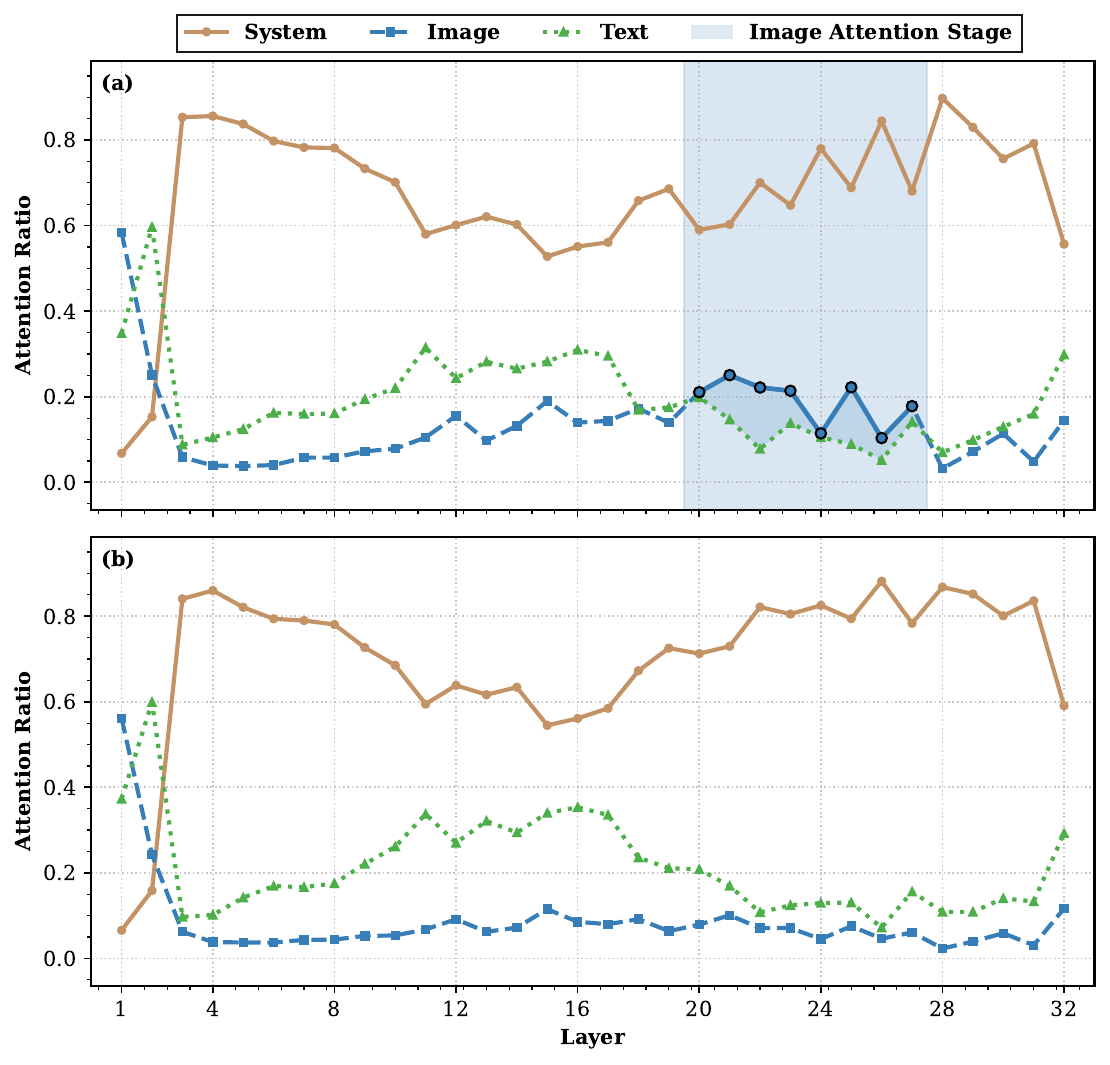}
  \caption{Attention Ratio across layers. \textbf{(a)} Object attention; \textbf{(b)} Non-object attention. Non-object tokens exhibit minimal attention to the image, while for object tokens we partition the attention into four stages according to their variations through layers, with the mid-to-late layers (20-27) being the Image-Attention Stage.}
  \label{fig:attn_layers}
\end{figure}
We first compare the attention distributions of generated object tokens versus non-object tokens, as shown in Fig.~\ref{fig:attn_layers}. It is observable that, apart from the initial 1-2 layers, non-object tokens exhibit minimal attention to the image. In contrast, object tokens exhibit a four-stage pattern:
\begin{itemize}
    \item Initialization Stage: In the first 1–2 layers, attention is high on both image and text, as the model rapidly aggregates multimodal context.
    \item Text-Attention Stage: In the early-to-mid layers (3-19), text attention dominates while image attention remains relatively low.
    \item \textbf{Image-Attention Stage}: In the mid-to-late layers (20-27), attention to the image increases and becomes dominant, while text attention drops. The model focuses on extracting evidence related to the candidate objects.
    \item Language-Organization Stage: In the final layers (28-32), image attention decreases again while text attention rises, as the model is shifting to organize and generate a grammatically and logically coherent output.
\end{itemize}

\begin{figure}[t]
  \centering
  \includegraphics[width=0.96\linewidth]{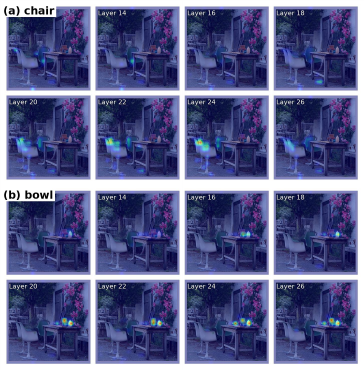}
  \caption{Both the real object token \textbf{(a)} `chair' and the hallucinated object token \textbf{(b)} `bowl' exhibit attention that converges on specific regions during the image-attention stage (layers 20–27).}
  \label{fig:real_hallu_focus}
\end{figure}

We then ask whether real and hallucinated object tokens differ in how much they attend to the image. We employ CHAIR~\cite{chair} to identify real and hallucinated object tokens and quantify their average image attention during the Image-Attention Stage, as shown in Fig.~\ref{fig:same_attention} (b). The results show no systematic difference between the two categories. To provide further qualitative evidence, we visualize the evolution of image attention across layers when generating object tokens. As shown in Fig.~\ref{fig:real_hallu_focus}, both real and hallucinated object tokens focus their attention on specific regions during the Image-Attention Stage, with comparable magnitudes. These observations indicate that \textbf{object hallucination is not caused by insufficient visual attention.}




\subsection{Different Truths: Semantic Inconsistency \\Revealed by Logit-Lens}
\label{subsec:different truths}
\begin{figure}[t]
  \centering
  \includegraphics[width=\linewidth]{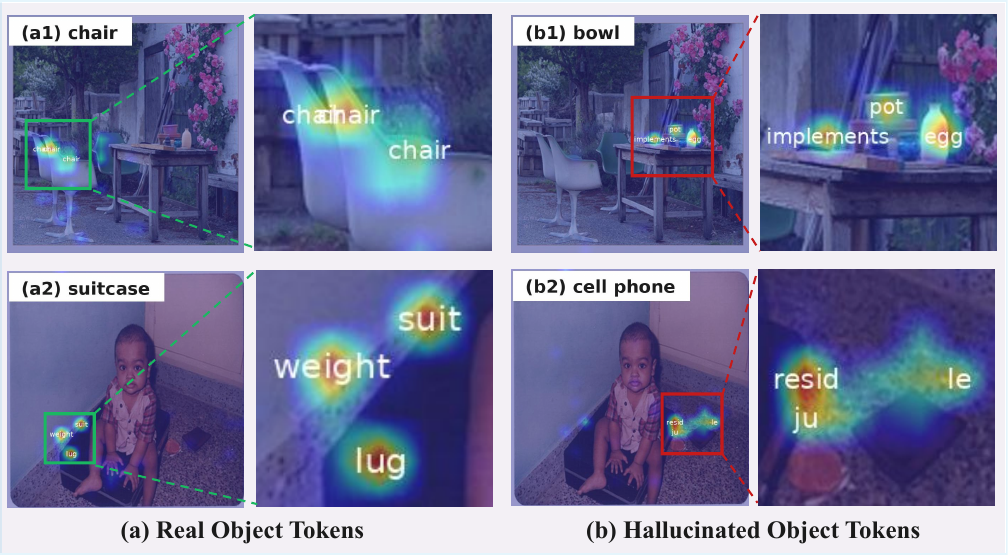}
  \caption{Via the Logit-Lens, semantics decoded from the high-attention regions in the Image-Attention Stage align with the \textbf{(a)} real object tokens, but diverge from \textbf{(b)} the hallucinated object tokens.}
  \label{fig:logit_lens}
\end{figure}

The previous section showed that 
hallucination is not related to \textbf{how much} the model attends to the image. In the following, we show its relation to \textbf{what} the model attends to. Specifically, we use Logit-Lens~\cite{logitlens} to ``read out'' the semantics of the attended visual regions. During the Image-Attention Stage, we project the hidden states of the top-$k$ attended image tokens into the vocabulary space, thereby decoding what the model semantically ``recognizes". The results are shown in Fig. ~\ref{fig:logit_lens}. We find:
\begin{itemize}
    \item For \textbf{Real Objects (Semantic Consistency)}: When the model generates a real object token (e.g., `suitcase'), the highly-attended visual regions (i.e., the actual suitcase regions) can also be decoded to consistent tokens like `suit' or `lug' by Logit-Lens. This indicates that the real token attention is faithfully grounded in the visual evidence.
    \item For \textbf{Hallucinated Objects (Semantic Inconsistency)}: When the model generates a hallucinated object (e.g., `cell phone'), a clear semantic inconsistency emerges. The highly-attended visual regions (e.g., a `blurry patch on the ground') do not decode to the hallucinated target token by Logit-Lens. This reveals a critical disconnect: the semantics ``read'' from the attended region are inconsistent with the final output.
\end{itemize}

This finding uncovers a key phenomenon of object hallucination: \textbf{hallucinated object tokens exhibit a semantic inconsistency, where the visual evidence extracted by the attention mechanism does not support the generated token.} This insight directly motivates our subsequent detection method, and empirical evidence for this phenomenon can be seen in the detection experiments in Sec.~\ref{subsec:hallucination detection}.

\subsection{Two Types:  Visual Uncertainty and Contextual Prior}
\label{subsec:two types}
The previous section showed that hallucinations can appear around the regions with strong attention during the Image-Attention Stage. In this section, we further investigate \textbf{why} the model attends to these regions by devising a High-Attention Regions Masking experiment.

\begin{figure}[t]
  \centering
  \includegraphics[width=\linewidth]{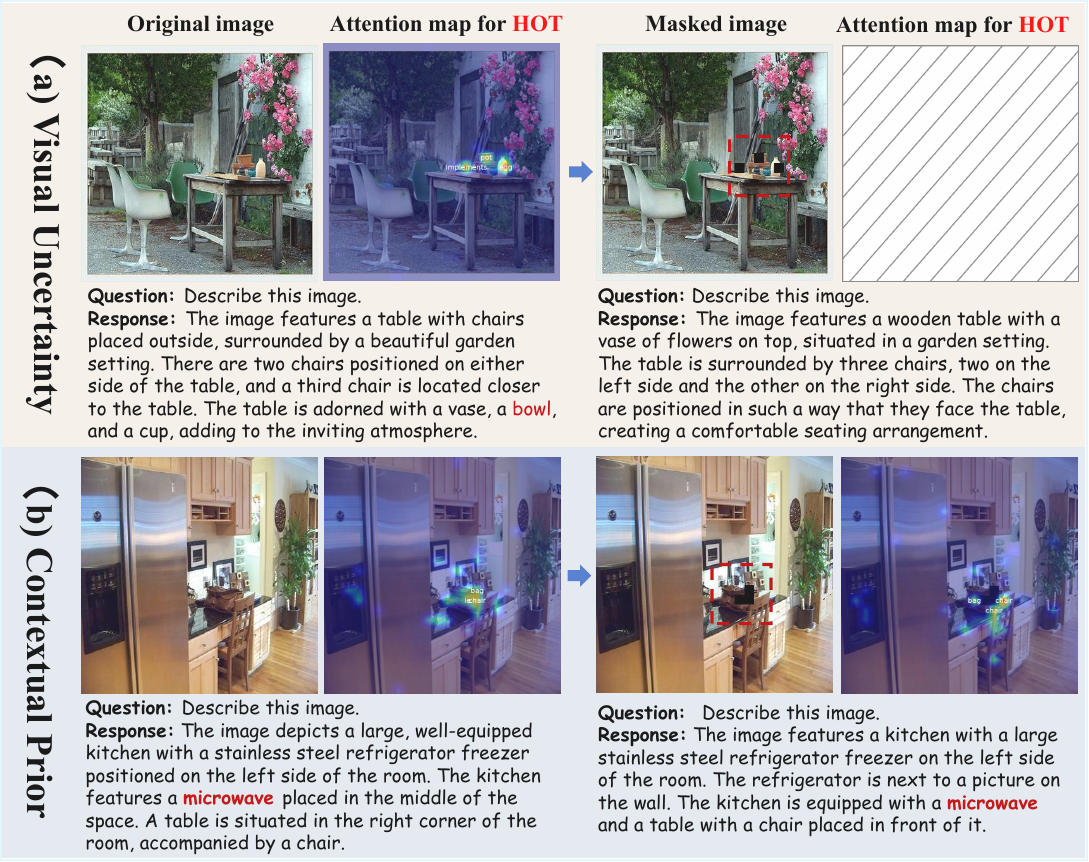}
  \caption{Two types of hallucination. \textbf{(a)} Disappears after masking high-attention regions, driven by visual uncertainty. \textbf{(b)} Persists after masking with attention shifting to other regions, driven by contextual prior. HOT here means Hallucinated Object Token.}
  \label{fig:two_type}
\end{figure}

Let the attention weight at layer $l$ on the $j$-th image token during the $t$-th generation step (targeting object token $o_t$) be $\alpha_{tj}^{(l)}$. Let $\mathcal{S}_{\mathrm{IA}}$ be the set of layers in the Image-Attention Stage. We first obtain the average attention that $o_t$ assigns to each image token $j$ in $\mathcal{S}_\mathrm{IA}$:
\begin{equation}
    \bar{\alpha}_{tj} = \frac{1}{|\mathcal{S}_{\mathrm{IA}}|} \sum_{l \in \mathcal{S}_{\mathrm{IA}}} \alpha_{tj}^{(l)}.
\end{equation}
We then select the top-\(k\) high-attention image tokens \(\Omega_t = \operatorname{TopK}(\bar{\alpha}_{t},k)\), mask corresponding patches in the image and regenerate the response to observe the model's behavior. As illustrated in Fig.~\ref{fig:two_type}, this experiment reveals two types of object hallucinations with distinct underlying causes:
\begin{itemize}
    \item \textbf{Visual Uncertainty Hallucination}: In these cases, the hallucinated token is directly related with high-attention regions. When we mask these regions, the hallucinated token disappears, indicating that the hallucination is visually rooted. Empirically, such regions are often ambiguous, blurry, or uncertain areas (e.g., a round, ceramic-like area leading to the generation of ``bowl''; or a dark, blurry rectangular region resulting in the output ``cell phone''). These observations suggest that the model may attempt to forcibly extract certain semantics from highly uncertain visual evidence, which is insufficient to support the generated object, leading to semantic inconsistency. Once this uncertain evidence source is removed, the model loses its visual anchor for generating hallucination.
    \item \textbf{Contextual Prior Hallucination}: In contrast, we also observe hallucination where masking the high-attention regions does not remove the hallucinated token. Instead, hallucination persists and attention shifts to other regions. Logit Lens analysis further shows that these attended regions can be decoded into the correct objects actually present in the image, yet the model still outputs the hallucinated token. This pattern is consistent with the generation being dominated by strong contextual priors \cite{pai} (e.g., outputting ``microwave'' in a kitchen description even when no microwave is present). In such cases, the attention mechanism is more like a procedural ``routine'': its internal rules require it to ``attend somewhere'' before generating an object. So it arbitrarily ``grounds'' its attention on some regions to satisfy this procedural requirement, while the actual generation decision is mainly controlled by the contextual prior.
\end{itemize}

We further perform a statistical analysis on the hallucination instances over 500 images. For LLaVA-1.5-7b, the ratio of ``Visual Uncertainty'' to ``Contextual Prior'' hallucination is approximately 2:1. The findings in this section expose two different types of object hallucination, demonstrating that a single, one-size-fits-all mitigation strategy is likely insufficient, thereby laying a solid foundation for our subsequent design of a targeted mitigating method.

\section{Detect-Mitigate Framework For Object Hallucination}
\begin{figure*}[t]
  \centering
  \includegraphics[width=\linewidth]{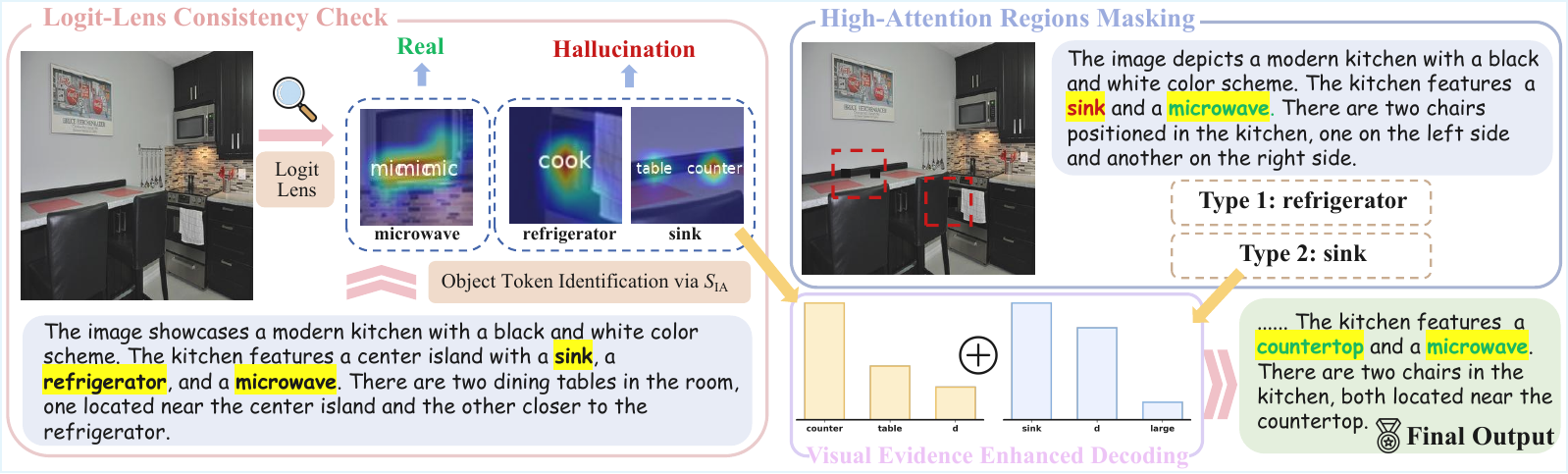}
  \caption{The overall structure of our Detect-Mitigate framework. It consists of three components. First, \textbf{Logit-Lens Consistency Check (LLCC)} detects hallucinated object tokens in the response. Then, \textbf{High-Attention Regions Masking (HARM)} categorizes them into two different types of hallucination while generating an intermediate output. If type 2 hallucination is detected, the output is further processed by \textbf{Visual Evidence Enhanced Decoding (VEED)} to produce the final result.}
  \label{fig:framework}
\end{figure*}

Based on the analysis in Sec.~\ref{sec:object}, we propose a simple yet effective Detect–Mitigate framework, as shown in Fig.~\ref{fig:framework}. We first introduce a Logit-Lens Consistency Check (LLCC) to detect potential hallucinated tokens in real time (motivated by Sec.~\ref{subsec:different truths}). We then determine which type (from Sec.~\ref{subsec:two types}) each detected hallucination belongs to and apply a targeted mitigation strategy. Details are as follows.

\subsection{Hallucination Detection via Logit-Lens Consistency Check}
\label{sec:hallucination detection}
Our detection module aims to identify object hallucination in real-time within the generated sequence. This process is based on the finding from Sec.~\ref{subsec:different truths}: real objects are semantically consistent with their visual evidence, whereas hallucinated objects are not. The detection process is divided into three stages: (1) Object Token Identification, (2) Semantic Consistency Check and (3) Hallucination Classification.

\subsubsection{Object Token Identification}
To precisely locate object tokens while avoiding unnecessary computation on non-object tokens, we design an identification step. First, inspired by the observation from Sec.~\ref{subsec:same attention} that object tokens exhibit significantly higher image attention than non-object tokens, we calculate the average image attention $\mathcal{A}_{img}(o_t)$ for each generated token $o_t$ within the Image-Attention Stage $\mathcal{S}_{\mathrm{IA}}$:
\begin{equation}
    \mathcal{A}_{img}(o_t) = \frac{1}{|\mathcal{S}_{\mathrm{IA}}|} \sum_{l \in \mathcal{S}_{\mathrm{IA}}} \left( \sum_{j=1}^{N_{img}} \alpha_{tj}^{(l)} \right),
\end{equation}
where $\alpha_{tj}^{(l)}$ is the attention weight of $o_t$ on the $j$-th image token at layer $l$, and $N_{img}$ is the total number of image tokens. We treat \(o_t\) with \(\mathcal{A}_{\mathrm{img}}(o_t) > \tau_{\mathrm{attn}}\) as object tokens, where $\tau_{\mathrm{attn}}$ is a threshold hyper-parameter.





\subsubsection{Semantic Consistency Check}
For each object token, we first collect its attention to all image tokens within $\mathcal{S}_{\mathrm{IA}}$ and select the Top-k attended tokens:
\begin{equation}
    \Omega_{t} = \operatorname{TopK}\big(\alpha_{tj}^{(l)},\,k\big), l \in \mathcal{S}_{\mathrm{IA}}.
\end{equation}
We apply Logit Lens to each hidden state $h(\omega_t^{i}), \omega_t^{i} \in \Omega_t$ and take the
highest-probability decoded token:
\begin{equation}
    v_t^{i} =
\arg\max_{v \in \mathcal{V}}
\bigl[\text{softmax}\bigl(W_U h(\omega_t^{i})\bigr)\bigr]_v,
i = 1,\dots,k .
\end{equation}
where $\mathcal{V}$ denotes the vocabulary. Then we compare whether $o_t$ is semantically consistent with each token $v_t^{i}$. Since $v_t^{i}$ might be a synonym of $o_t$ (e.g., $o_t$=car, $v_t^{i}$=vehicle), we adopt a standard similar to~\cite{chair,amber}, using WordNet~\cite{wordnet} or another semantic similarity function $\text{Sim}(\cdot, \cdot)$:
\begin{equation}
    \text{Label}(o_t) =
    \begin{cases}
        0 \ (\text{Real}) , & \text{if } \text{Sim}(o_t, v_t^{i}) > \tau_{\mathrm{sim}}, \\
        1 \ (\text{Hallucination}), & \text{otherwise}.
    \end{cases}
\end{equation}
where $\tau_{sim}$ is the semantic similarity threshold. 

\subsubsection{Hallucination Classification}
When a token $o_t$ is detected as ``hallucination'', we further classify its type. Following the same procedure as Sec.~\ref{subsec:two types}, we mask the high-attention regions within $\mathcal{S}_{\mathrm{IA}}$ and observe whether the hallucination ``disappears''. Let the newly generated output sequence be $O_{new}$, then:
\begin{equation}
    \text{Type}(o_t) =
    \begin{cases}
        1 \ (\text{Visual Uncertainty}) & \text{if } o_t \notin O_{\mathrm{new}}, \\
        2 \ (\text{Contextual Prior})   & \text{if } o_t \in O_{\mathrm{new}}.
    \end{cases}
\end{equation}

\subsection{Targeted Mitigation Methods for Hallucination}
\label{sec:hallucination mitigation}
A single strategy cannot effectively handle hallucinations arising from different mechanisms. We therefore propose two simple yet effective methods tailored to each class.
\subsubsection{Visual Uncertainty: High-Attention Region Masking}
As analyzed in Sec.~\ref{subsec:two types}, this type of hallucination occurs when the model anchors its attention on a high-uncertainty visual region, and the generation heavily depends on this erroneous evidence. Therefore, we employ a simple masking strategy, identical to that in Sec.~\ref{subsec:two types}.

Let all the Top-k high-attention image tokens of each hallucinated object token be $\Omega$ and the original image be $I$. Its binary mask $M$ satisfies:
\begin{equation}
    \mathbf{M} = \begin{cases} 1, & \text{if pixel/patch } p \in \Omega \\ 0, & \text{otherwise} \end{cases}.
\end{equation}
Let $\mu$ be a simple replacement value, e.g., the mean image color or a zero vector (black). The masked image $I^{\mathrm{mask}}$ is:
\begin{equation}
    I^{\mathrm{mask}} = (1-\mathbf{M})\odot I + \mathbf{M}\odot \mu.
\end{equation}
We then regenerate with the same prompt using $I^{\mathrm{mask}}$. By removing the high-uncertainty evidence, this strategy effectively reduces hallucination of this type.

\subsubsection{Contextual Prior: Visual Evidence Enhanced Decoding}
The second type is insensitive to masking. Its root cause is that strong contextual priors dominate the output, while the real visual evidence is ignored. To address this, we propose Visual Evidence Enhanced Decoding (VEED). During the decoding stage of the language model, we directly inject visual semantics from high-attention regions to increase their weight in the final logits, thereby exploiting correct visual evidence and suppressing the influence of the priors.

Let $h(\omega_t^{\mathrm{max}})$ denote the hidden state of the most attended visual region at step $t$. We extract its visual logits using Logit Lens (computed in Sec.~\ref{sec:hallucination detection}): $z^{\mathrm{vis}}_t = W_U h(\omega_t^{max})$.
Let $z^{\mathrm{mask}}_t = \mathrm{logits}_\theta(y_t \mid y_{<t}, I^{\mathrm{mask}})$
be the model’s decoding logits at step $t$ when the high-attention region is masked.
We fuse them to form the enhanced distribution for type 2:
\begin{align}
    p_t &= \operatorname{softmax}\big((1-\alpha) \cdot z_t^{\mathrm{vis}}+\alpha \cdot z_t^{\mathrm{mask}}\big) \\
    &= \operatorname{softmax}\Big( (1-\alpha) \cdot W_U h(\omega_t^{\mathrm{max}}) \nonumber \\
    & \qquad + \alpha \cdot \mathrm{logits}_\theta(y_t \mid y_{<t}, I^{\mathrm{mask}}) \Big) \nonumber
\end{align}

Here, $\alpha$ controls the strength of visual-evidence injection: a lower value gives higher weight to visual semantics from the high-attention region in the final decision.

\section{Experiments}
In this section, we empirically validate the effectiveness of our proposed Detect–Mitigate framework. We first assess the hallucination detection method in Sec.~\ref{sec:hallucination detection}, and then the mitigation method in Sec.~\ref{sec:hallucination mitigation}.

\subsection{Hallucination Detection}
\label{subsec:hallucination detection}
\subsubsection{Experimental Settings}
Following~\cite{opera,devils}, we randomly sample 500 images from COCO2014 dataset, generate descriptions with LLaVA-1.5-7B, and use CHAIR~\cite{chair} to label hallucinated and real objects. We compare our approach against three recent methods: Uncertainty Score~\cite{us}, which identifies hallucinations via uncertainty estimates; InterConf~\cite{interconf}, which aggregates decoded probabilities from all layers; and SVAR~\cite{devils}, which exploits attention weights from intermediate layers 5–18. We set $\tau_{attn}$ to 0.15 and $k$ to 3. Following ~\cite{amber}, we set $\tau_{sim}$ to 0.8 to match semantically similar words. Additional parameter experiments are in the appendix.

\subsubsection{Results and Analysis}
As shown in Tab.~\ref{tab:detection}, LLCC achieves the best Precision, Recall, and F1. By explicitly localizing object-token attention and performing a semantic consistency check on the attended regions, it directly tests whether the generation is supported by visual evidence. In contrast, baselines rely on decoding probabilities or cross-layer aggregates, weakening their causal link to the image. Uncertainty Score measures uncertainty based on probability, which lacks a theoretical link between probability and uncertainty and is undermined by LLM's overconfidence~\cite{overconfidence}. InterConf merely assesses whether the object exists in the image, lacking explicit localization, so language priors or misleading visual cues can make it wrong, and its reliance on all layers further degrades accuracy. SVAR focuses on attention ``quantity" (Summed Ratio) over ``quality". Our analysis (Sec.~\ref{subsec:same attention} / Sec.~\ref{subsec:different truths}) shows that attention aligns with image evidence primarily in the later Image-Attention Stage, while it aggregates signals from non-image-dominant layers, which introduces noise and dilutes the signal. Our findings (Sec.~\ref{subsec:two types}) also suggest quantity maybe not a very reliable metric. While for our method, it uniquely assesses the ``reasonableness of the generation" by first locking onto the visual source for object tokens and then checking if that source provides consistent semantic support. Notably, it can also identify different causal types of hallucinations, which prior methods cannot.

\begin{table}[t]
\small
  \centering
  \caption{Hallucination Detection Results. Best results are in \textbf{bold}, and second-best are \underline{underlined}.}
  \label{tab:detection}
  \resizebox{0.44\textwidth}{!}{
  \begin{tabular}{l|c|c|c}
    \toprule
    \multirow{1}{*}{Method} 
    & \multicolumn{1}{c|}{Precision} 
    & \multicolumn{1}{c|}{Recall} 
    & \multicolumn{1}{c}{F1 Score} \\
    \midrule
    Uncertainty Score & 0.5965 & 0.6415 & 0.6182 \\
    InterConf & \underline{0.6717} & 0.6907 & 0.6811 \\
    SVAR & 0.6500 & \underline{0.7222} & \underline{0.6842} \\
    \textbf{LLCC(Ours)} & \textbf{0.7870} & \textbf{0.7955} & \textbf{0.7932} \\
    \bottomrule
  \end{tabular}}
\end{table}

\begin{table*}[htbp]
\small
  \centering
  \caption{CHAIR hallucination evaluation results on multiple LVLMs. Best results are in \textbf{bold}, and second-best are \underline{underlined}. DeCo and Devils do not report analyses on Qwen2-VL so their experimental settings are unspecified. We mark them as ``–”.}
  \label{tab:chair}
  \resizebox{1.0\linewidth}{!}{
  \begin{tabular}{l|cc|cc|cc|cc}
    \toprule
    \multirow{2}{*}{\textbf{Method}} 
    & \multicolumn{2}{c|}{\textbf{LLaVA-1.5-7B}} 
    & \multicolumn{2}{c|}{\textbf{LLaVA-1.5-13B}} 
    & \multicolumn{2}{c|}{\textbf{Shikra-7B}} 
    & \multicolumn{2}{c}{\textbf{Qwen2-VL-7B}} \\
    \cmidrule(lr){2-3} \cmidrule(lr){4-5} \cmidrule(lr){6-7} \cmidrule(lr){8-9}
    & $CHAIR_S$ $\downarrow$ & $CHAIR_I$ $\downarrow$ & $CHAIR_S$ $\downarrow$ & $CHAIR_I$ $\downarrow$ & $CHAIR_S$ $\downarrow$ & $CHAIR_I$ $\downarrow$ & $CHAIR_S$ & $\downarrow$ $CHAIR_I$ $\downarrow$ \\
    \midrule
    Greedy & 49.8 & 20.4 & 47.8 & 19.8 & 58.4 & 22.2 & 31.4 & 12.7 \\
    Beam & 50.0 & 20.1 & 48.2 & 18.9 & 59.7 & 23.1 & 30.8 & 11.9 \\
    Nucleus & 59.0 & 25.0 & 54.2 & 21.7 & 58.0 & 22.0 & 33.3 & 14.0 \\
    VCD & 56.3 & 22.9 & 50.3 & 18.9 & 52.4 & 19.8 & 32.1 & 13.2 \\
    OPERA & 42.9 & 18.7 & 42.1 & 16.4 & 38.1 & 16.7 & 28.0 & 10.3 \\
    DeCo & 37.3 & 16.5 & 38.3 & 15.7 & 40.2 & 16.9 & - & - \\
    Devils & 32.1 & 13.7 & 35.4& 14.1 & \underline{32.8} & \underline{13.1} & - & - \\
    PAI & \underline{29.8} & \underline{13.2} & \underline{33.2} & \underline{13.5} & 37.9 & 15.0 & \underline{24.7} & \underline{8.6} \\
    \textbf{Ours} & \textbf{26.8} & \textbf{10.0} & \textbf{31.3} & \textbf{12.4} & \textbf{31.4} & \textbf{12.7} & \textbf{24.0} & \textbf{8.3} \\
    
    \bottomrule
  \end{tabular}}
\end{table*}

\begin{table*}[htbp]
\small
  \centering
  \caption{AMBER hallucination evaluation results. Best results are in \textbf{bold}, and second-best are \underline{underlined}.}
  \label{tab:amber}
  \resizebox{1.0\linewidth}{!}{
  \begin{tabular}{l|cccc|cccc|cccc}
    \toprule
    \multirow{2}{*}{\textbf{Method}} 
    & \multicolumn{4}{c|}{\textbf{LLaVA-1.5-7B}} 
    & \multicolumn{4}{c|}{\textbf{LLaVA-1.5-13B}} 
    & \multicolumn{4}{c}{\textbf{Shikra-7B}} \\
    \cmidrule(lr){2-5} \cmidrule(lr){6-9} \cmidrule(lr){10-13}
    & $CHAIR$ $\downarrow$ & $Cover$ $\uparrow$ & $Hal$ $\downarrow$ & $Cog$ $\downarrow$ & $CHAIR$ $\downarrow$ & $Cover$ $\uparrow$ & $Hal$ $\downarrow$ & $Cog$ $\downarrow$ & $CHAIR$ $\downarrow$ & $Cover$ $\uparrow$ & $Hal$ $\downarrow$ & $Cog$ $\downarrow$ \\
    \midrule
    Greedy & 6.9 & 51.0 & 32.0 & 3.3 & 6.8 & 52.0 & 31.7 & 3.5 & 10.6 & 52.0 & 47.0 & 5.2\\
    Beam & 7.8 & 49.7 & 35.5 & 4.1 & 8.7 & 48.7 & 38.8 & 3.8 & 10.0 & 51.3 & 44.1 & 4.8\\
    Nucleus & 7.9 & 51.1 & 35.5 & 4.3 & 8.4 & 51.5 & 38.1 & 3.9 & 11.1 & 52.2 & 49.4 & 5.7\\
    VCD & 6.4 & \textbf{52.1} & 33.2 & 2.9 & 7.6 & \textbf{53.2} & 35.3 & 3.4 & 10.3 & \textbf{53.7} & 45.2 & 5.0\\
    OPERA & 6.8 & 47.3 & 28.3 & 2.7 & 6.9 & 49.0 & 31.5 & 3.3 & 8.2 & 51.6 & 39.8 & 3.9\\
    DeCo & 6.6 & 46.4 & 27.6 & 2.8 & 6.2 & 48.5 & 27.7 & \underline{2.9} & 7.5 & 48.2 & 37.9 & 4.0\\
    Devils & \underline{3.5} & 50.2 & 19.9 & \underline{1.3} & 5.6 & 51.1 & 30.6 & 3.4 & 6.3 & 51.4 & \underline{32.1} & \underline{2.8} \\
    PAI & 4.2 & 50.7 & \underline{18.4} & 1.6 & \underline{4.9} & 51.9 & \underline{26.2} & 3.0 & \underline{5.9} & 50.9 & 32.7 & 3.4\\
    \textbf{Ours} & \textbf{2.8} & \underline{51.2} & \textbf{14.7} & \textbf{1.2} & \textbf{4.0} & \underline{52.0} & \textbf{24.1} & \textbf{2.5} & \textbf{5.3} & \underline{52.3} & \textbf{30.3} & \textbf{2.0}\\
    
    \bottomrule
  \end{tabular}}
\end{table*}

\subsection{Hallucination Mitigation}
\subsubsection{Experimental Settings}
\textbf{Benchmark and metrics.} We evaluate on two standard benchmarks: \textbf{CHAIR}~\cite{chair} and \textbf{AMBER}~\cite{amber}. \textbf{CHAIR} measures whether objects mentioned in the generated text appear in the image and reports sentence-level and instance-level metrics ($\mathrm{CHAIR}_S$ and $\mathrm{CHAIR}_I$, lower is better):
\[
\resizebox{\columnwidth}{!}{$
\mathrm{CHAIR}_I=
\frac{\lvert\text{hallucinated objects}\rvert}
     {\lvert\text{all objects mentioned}\rvert},
\mathrm{CHAIR}_S=
\frac{\lvert\text{sentences with hallucinated object}\rvert}
     {\lvert\text{all sentences}\rvert}
$}
\]
Following common settings~\cite{opera,pai,devils}, we randomly sample 500 images from the COCO2014 validation set for CHAIR evaluation. \textbf{AMBER} is a multi-dimensional benchmark comprising 1004 images with detailed object annotations. We report four metrics under its generative setting: CHAIR (rate of hallucinated objects), Cover (object coverage), Hal (fraction of responses containing any hallucination), and Cog (tendency to produce cognitively tempting target objects), where higher is better for Cover and lower is better for the others. \\ 
\textbf{Baselines.} We compare against three common decoding strategies and five representative hallucination mitigation methods. Greedy Search selects the most probable next token at each step. Beam Search maintains multiple hypotheses and outputs the highest-probability sequence. Nucleus Sampling samples from tokens whose cumulative probability exceeds a threshold $p$. VCD~\cite{vcd} contrasts outputs generated from the original and a perturbed image to suppress language priors. OPERA~\cite{opera} introduces dynamic penalties on over-confident decoding steps. DeCo~\cite{deco} selects an intermediate ``anchor” layer to correct final-layer logits. Devils~\cite{devils} exploits mid-layer visual signals by aligning multi-head attentions. And PAI~\cite{pai} amplifies the attention of image tokens and contrasts its output with no-image logits.\\ 
\textbf{Implementation details.} We implement our method with greedy decoding as a representative case and test on three VLMs: LLaVA-1.5~\cite{llava1.5}, Shikra~\cite{shikra} and Qwen2-VL~\cite{Qwen2VL}. For Beam search, we set \(n_{beam}=3\). 
For Nucleus sampling, we set top-\(p=0.9\) and temperature \(\tau=0.7\). Following the original papers, we pair VCD with nucleus sampling, apply OPERA and DeCo on top of beam search, and combine Devils and PAI with greedy decoding. All experiments are conducted on a single NVIDIA A100 40GB GPU.
    

\subsubsection{Results and Analysis}
As shown in Tab.~\ref{tab:chair} and Tab.~\ref{tab:amber}, our method consistently reduces hallucination while maintaining (even slightly improving) object coverage. On CHAIR, we achieve the largest drop among the four models. On AMBER, we obtain the lowest hallucination rates across models while keeping strong coverage, which is slightly higher than the original 51.0 and second only to VCD’s 52.1 on LLaVA-1.5-7B. Notably, most baselines decrease on Cover rate, indicating that they sacrifice valid content to mitigate hallucination, whereas our approach achieves the lowest hallucination rates without eroding the original expression. We attribute the advantage to our two-stage framework: (i) explicit object-token localization and cause-aware hallucination detection, and (ii) targeted mitigation: masking only a minimal set of image tokens and applying decoding enhancement only for the second hallucination type, thereby limiting impact on original content. Although absolute scores vary with model scale and architecture, the relative ranking remains stable, demonstrating robustness of our method. Additional ablations are provided in the appendix.


\section{Conclusion}
In this work, we revisit object hallucination in LVLMs via the lens of attention and Logit Lens analysis. By analyzing object token generation process, we reveal a ``same attention, different truths'' phenomenon: both real and hallucinated objects exhibit similar attention patterns in the image-attention stage, yet only the former are semantically decodable from the attended regions via Logits-Lens.  
Building on this, we propose a training-free Detect–Mitigate framework that first detects hallucinated object tokens via Logit-Lens Consistency Check and then applies cause-aware High Attention Regions Masking for visual uncertainty hallucination and Visual Evidence Enhanced Decoding for contextual prior hallucination.  
Extensive experiments demonstrate the effectiveness and generality of our method.

{
    \small
    \bibliographystyle{ieeenat_fullname}
    \bibliography{main}
}

\end{document}